\documentclass[9pt,journal]{IEEEtran}
\usepackage{amsmath,amssymb,amsfonts}
\usepackage{graphicx}
\usepackage{textcomp}
\usepackage{color}
\pdfoutput=1
\begin{document}
\title{Haptic Teleoperation of UAVs through Control Barrier Functions}
\author{Dawei Zhang, \IEEEmembership{Student Member, IEEE}, Guang Yang, \IEEEmembership{Student Member, IEEE}, Rebecca P. Khurshid, \IEEEmembership{Member, IEEE}
\thanks{Dawei Zhang is with the Department of Mechanical Engineering, Boston University, Boston, MA 02215 USA (e-mail: dwzhang@bu.edu)}
\thanks{Guang Yang is with the Division of System Engineering, Boston University, Boston, MA 02215 USA(e-mail: gyang101@bu.edu).}
\thanks{Rebecca P. Khurshid is with the Department of Mechanical Engineering and the Division of Systems Engineering, Boston University, Boston, MA 02215 USA (e-mail: khurshid@bu.edu).}}

\maketitle

\begin{abstract}
This paper presents a novel approach to haptic teleoperation. Specifically, we use control barrier functions (CBFs) to generate force feedback to help human operators safely fly quadrotor UAVs. CBFs take a control signal as input and output a control signal that is as close as possible to the initial control signal, while also meeting specified safety constraints. In the proposed method, we generate haptic force feedback based on the difference between a command issued by the human operator and the safe command returned by a CBF. In this way, if the user issues an unsafe control command, the haptic feedback will help guide the user towards the safe input command that is closest to their current command. We conducted a within-subject user study, in which 12 participants flew a simulated UAV in a virtual hallway environment. Participants completed the task with our proposed CBF-based haptic feedback, no haptic feedback, and haptic feedback generated via parametric risk fields, which is a state-of-the-art method described in the literature. The results of this study show that CBF-based haptic feedback can improve a human operator's ability to safely fly a UAV and reduce the operator's perceived workload, without sacrificing task efficiency.

\end{abstract}


\section{Introduction}
\label{sec:introduction}

When remotely controlling an unmanned areal vehicle (UAV), the human operator typically needs to perceive the remote environment using only two-dimensional visual feedback. The limited field of view often leads to low levels of situational awareness, which can make it difficult to safely and accurately control the UAV \cite{mccarley2005human, Brandt2010}. Providing the human operator with force-based haptic feedback about the robot's environment has proven to help reduce collisions between a human-controlled UAV and the environment and improve operator situational awareness \cite{Brandt2010, Hou2013}. Although haptic feedback helps to reduce the number of collisions between the UAV and the environment, experimental evaluation of methods that use impedance-type force-feedback devices shows that the rate of collisions still remains relatively high even with haptic feedback \cite{Brandt2010, Hou2013}. 

In this paper, we present the design and evaluation of a new force-feedback method based on control barrier functions (CBFs), which have recently become popular in the domain of safety-critical control. As discussed in Section \ref{sec:background}, CBFs act as an intermediate layer between desired control inputs and actual control inputs sent to the robot, ensuring that the actual control signals sent to the robot are both safe and as close to the desired control input as possible. In this paper, we use a CBF to find a safe control input that is closest to a human operator's desired control input. We then used the difference between the control input provided by the human operator and the safe input returned by the CBF to generate force feedback for the user. \textbf{This force feedback, therefore, helps guide the human user towards the input command that is closest to their current command and deemed to be safe.} This contrasts with previous methods that provide force feedback in the opposite direction of the obstacle that is deemed to pose the highest risk of collision. 

%CBFs are well-suited for used in human-controlled robotic applications because CBFs can be used to minimize the difference between the desired control input provided by the human operator and the actual control input sent to the robot. Thus ensuring that the robot 

%offer methods to ensure robot safety by dictating that all control inputs meet some safety constraints (Citation). For example, control barrier functions (CBFs) can be used as an intermediate layer between desired control inputs and actual control inputs sent to the robot. CBFs guarantee that the actual control sent to the robot will never cause the robot to take an unsafe action, such as colliding with a wall. 

%control barrier functions ensure that the robot remains safe at all times, but only allows control inputs that meet certain safety constrains, such as ensuring the robot does not collide with an obstacle in the environment.

%However, past haptic feedback schemes do not take into account the user's commands, instead only use the state of the UAV to generate haptic feedback. Our key insight is that the user's commands should be considered when designing haptic feedback.} In this paper we leverage control barrier functions (CBF) to generate haptic feedback based on both the UAV's state and the operator's commands. 

The remaining of the paper is organized as follows. A related background is described in Section \ref{sec:background}. A detailed description of our method is presented in Section \ref{sec:methods}. Section \ref{sec:user study} describes the design of a user study investigating the effect of the CBF-generated haptic feedback on the teleoperation of a simulated UAV, as compared to no haptic feedback and haptic feedback provided by another state-of-the-art method described in the literature. Results of the user study, and our interpretation of these results, is presented in Sections \ref{sec:results} and \ref{sec:discussion}, respectively. Finally, Section \ref{sec:conclusion} presents the main conclusions of this paper and our plans for future research.

\section{Background}
\label{sec:background}
\subsection{Haptic Feedback and Guidance in Teleoperation}
Haptic feedback is known to be beneficial in helping human operators remotely control a robot \cite{okamura2009haptic,pacchierotti2015cutaneous}. Many haptic feedback systems seek to improve the transparency of the teleoperation system by allowing the human to feel what the remote robot feels \cite{Niemeyer2008}.

Haptic feedback systems have also been designed to influence the operator's actions to enable them to better control the robot. Such feedback can either help guide the user along a desired trajectory \cite{Omari2013} or can help the user keep the robot away from restricted regions \cite{Brandt2010,Hou2013,Lam2009}. Helping keep the robot away from restricted regions has been shown to reduce collisions and increase situational awareness in the control of UAVs \cite{Hou2013}. Furthermore, keeping the robot away from restricted areas does not require prior knowledge or online predictions of the task goal. Therefore, our work focuses on the design of haptic feedback to keep the robot away from restricted regions. 

The use of virtual fixtures is one popular method to help keep the robot away from unsafe areas \cite{abbott2007haptic,bowyer2013active}. Virtual fixtures are binary in nature and only exert forces when the robot is physically in contact with the restricted region, much as a ruler only exerts force on a pencil when the pencil is contacting the ruler. While virtual fixtures are useful in some domains, such as certain types of robot-assisted surgery \cite{okamura2004methods}, there are many situations where the robot should not physically contact the restricted region. For example, a UAV that comes into contact with the wall would be unsafe, posing a risk to both the UAV and its environment. Therefore, other researchers have created haptic feedback to serve as a warning about the location of restricted regions in the environment.

%For the aim of safety and in pursuit of telepresence, haptic feedback has been ubiquitously used in the teleoperation of manipulators \cite{park2006haptic} and telerobotic surgery systems \cite{okamura2004methods}, \cite{okamura2009haptic}. Haptic feedback also plays as a platform that leverages the mature skills of human operators and the advanced robotic technologies \cite{Niemeyer2008}.

%\textcolor{red}{Most haptics allows the human operator to feel what the robot feels. Some are used to help guide the operator's actions.} For example, virtual fixtures use forces to help keep the robot away from unsafe areas and/or can help the user follow a desired path with the robot \cite{abbott2007haptic,bowyer2013active}. Virtual fixtures to help keep the robot away from unsafe areas are typically binary in nature and only exert forces when the robot is commanded to the unsafe area. 

%\textcolor{red}{Similar to virtual fixtures, some researchers use haptic feedback to assist the operator to navigate the robot.} For instance, Vander et al. improved the performance of navigational assistance through narrow passages by applying assisted haptic feedback to the powered wheelchair \cite{vander2012powered}. In a similar way, Lee et al. applied both environmental force and collision-preventing force in the teleoperation of the ground mobile vehicles. They indicted that the force provided to the user can help them navigate more safely and have a better cognition of the remote environment \cite{lee2005haptic}.

Haptic feedback that warns the human operator of a risk of collision is particularly relevant to the teleoperation of UAVs \cite{Brandt2010,Lam2009}. A grounded kinesthetic (force-reflecting) control interface can be used to apply a force on the user when there is an increased risk of a collision. In these systems, the magnitude of the force is related to the risk of a collision and the direction of the force points directly away from the object that poses the greatest risk. For example, drawing on potential functions used in robotic path planning, Lam et al. proposed a parametric risk field (PRF) to calculate the risk of a collision \cite{Lam2009}.  In another example, Brant and Colton set the magnitude of the force of the haptic feedback to be proportional to the time that it would take the UAV to collide with an object in its environment if the UAV continued flying with its current velocity \cite{Brandt2010}. The results of a user study showed that this time-to-impact method was effective in reducing the number of collisions between the robot and its environment, without sacrificing task efficiency \cite{Brandt2010}. However, the rate of collisions remained relatively high. Hou and Mahony implemented both of the above methods and found that participants crashed a UAV at similar rates under both methods.
%\textcolor{red}{ A limitation of this method is that numerous parameters must be tuned for this method. Furthermore, these parameters will likely have to be tuned for different UAV environments and task goals.}

We note, that instead of giving 'warning' to the human operator, Hou and Mahony physically prevent the operator from issuing a command that would result in a collision by employing an admittance-type haptic device \cite{Hou2013}. Although this is a valid approach, taking the control authority away from the human operator can be undesirable. 

\subsection{Control Barrier Functions} 

Much like a physical barrier prevents physical objects from entering restricted regions, barrier functions (BFs) were first implemented in the optimization literature to prevent solutions being returned from undesirable regions \cite{prajna2004safety,prajna2006barrier,Ames2019}. Ames et al. extended the notion of BFs to control barrier functions (CBFs) by adding safety constraints to optimization-based control methods \cite{Ames2014}. Here CBFs are used to ensure that, if the robot is in a safe state (i.e. not in collision with an obstacle), the control input applied to the robot will keep the robot in a safe state. Wang et al. extended the use of CBFs beyond optimal control methods by creating a framework that can enforce safety constraints given control input generated by any method \cite{Wang2017}. This formulation of CBFs can be represented as:  

\begin{equation}\label{eqn:opimization}
\begin{array} { c l } { \vec{u}  = \underset { \vec{u} \in \mathbb { R } ^ { m } } { \operatorname { argmin } } } & | \vec{u} - \vec{u}_{ref}|   \\  \text { s.t. }  &  \text { safety constraints},   \end{array}
\end{equation}
  
where $\vec{u}_{ref}$ is the control input generated by any method, $\vec{u}$ is the safe control input that is sent to the robot, and $| \vec{u} - \vec{u}_{ref}|$ is some metric of the distance between $\vec{u}_{ref}$  and $\vec{u}$. In this formulation, CBFs can be thought of as a function that takes any control signal $\vec{u}_{ref}$ as an input and returns a control signal $\vec{u}$, which is the control signal that is as close as possible to $\vec{u}_{ref}$, while satisfying some safety criteria. 

Typically, CBFs are used to enforce safety constraints on reference control signals generated by autonomous methods. However, CBFs are also well-suited to enforce safety in human-controlled robot applications. By taking  $\vec{u}_{ref}$ to be a control signal generated from human commands, Xu and Sreenath used a CBF to achieve safe teleoperation of UAV quadrotors, ensuring that the UAV always remained in a safe physical region \cite{Xu2018}. A potential limitation of this work is that the method alters the control signal generated by the human user, which may reduce the human's understanding of the mapping between their actions and the robot's actions.

In this paper, we still enable the human operator to maintain full control of the UAV and always use the human-generated control signal as the control signal sent to the UAV. However, we seek to improve the human's ability to safely control a UAV by using CBFs to generate haptic feedback to help guide the user to a safe control input.

\section{HAPTIC FEEDBACK DESIGN}\label{sec:methods}

%Both \cite{Nguyen2016} and \cite{Xiao2019} introduced the CBFs to deal with the systems with high relative-degree constraints. In this paper, we adopt the high order control barrier functions (HOCBFs) to formulate our haptic teleoperation system.

%In this paper, we consider quadrotor UAV systems in which the speed of rotation of the rotors, $\omega$, is related to the thrust applied to the UAV, $T$ through the equation:    
%\begin{equation}
%T =  k_{T}\sum_{i=1}^{4} \omega_{i}^{2}    
%\end{equation} where $k_{T}$ is a constant parameter.

%The acceleration of the UAV in a Cartesian reference frame with the Z axis pointing up, therefore takes the form 
%\begin{equation}
%a = \frac{T}{m}R\left(\begin{array}{c}{0} \\ {0} \\ {1}\end{array}\right) - g \left(\begin{array}{c}{0} \\ {0} \\ {1}\end{array}\right),
%\end{equation}
%where $g$ is the acceleration due to gravity, $R$ is the rotation matrix representing the orientation of the UAV, $m$ is the mass of the UAV. 

%Because the commands sent to 

%The dynamics of a UAV can be modeled as an affine control system of the form:
%\begin{equation}
%\dot { \vec{x} } = f ( \vec{x} ) + g ( \vec{x} ) \vec{u}
%\end{equation}

%If describing the motion of the UAV is Cartesian space, then \vec{x} 

In this paper, we consider quadrotor UAVs. A common representation of the pose of quadrotor UAVs is given by $\sigma = [x, y, z, \psi]$, where $x$, $y$, and $z$ are the Cartesian components of the UAV's center of mass and $\psi$ is the yaw angle of the UAV \cite{mellinger2011minimum}. We choose to approximate the shape of the UAV as a horizontal disk, so that only the quadrotor's $x$, $y$, and $z$ positions will determine whether it crashes with other objects in the environment. Thus, our proposed haptic feedback only relates to the human operator's command of the position of the quadrotor's center of mass. No haptic feedback is provided related to the user's command of $\psi$.

%in which the speed of rotation of the rotors is related to the thrust applied to the UAV. Therefore, the control input (i.e. speed of rotation of the rotors) is proportional to forces and torques experienced by UAV's body, which map directly to the UAV's acceleration (citations). 

For quadrotor UAVs, the speed of rotation of the rotors is related to the thrust applied to the UAV. Therefore, the control input (i.e. speed of rotation of the rotors) is proportional to forces and torques experienced by UAV's body, which map directly to the UAV's acceleration \cite{voos2009nonlinear}. Therefore, the dynamics of the UAV can be modeled as a second-order integrator, in which the control input $u$ can be considered as the acceleration command of the UAV. For a second-order integrator, given a control input $\vec{u}$ at time $t$, the velocity and position of the UAV at the next time step, $t + dt$ are estimated to be: 
\begin{equation} \label{eq:double1}
\vec{v}_{t+dt} = \vec{v}_{UAV,t} + \vec{u}dt.
\end{equation} 

\begin{equation} \label{eq:double2}
\vec{x}_{t+dt} = \vec{x}_{UAV,t} + \vec{v}_{UAV,t}dt + \frac{1}{2}\vec{u}dt^2.
\end{equation}
where $\vec{v}_{UAV,t}$ denotes the current velocity of the UAV, $dt$ is the rate of the control loop.

%\textcolor{blue}{
%The rotation matrix that represents the rtansformation from the body fixed frame to the world frame can be expressed as follows:
%\begin{equation}
%\boldsymbol{R}=\left(\begin{array}{ccc}{c_{\psi} c_{\theta}} & {c_{\psi} s_{\theta} s_{\phi}-s_{\psi} c_{\phi}} & {c_{\psi} s_{\theta} c_{\phi}+s_{\psi} s_{\phi}} \\ {s_{\psi} c_{\theta}} & {s_{\psi} s_{\theta} s_{\phi}+c_{\psi} c_{\phi}} & {s_{\psi} s_{\theta} c_{\phi}-c_{\psi} s_{\phi}} \\ {-s_{\theta}} & {c_{\theta} s_{\phi}} & {c_{\theta} c_{\phi}}\end{array}\right)
%\end{equation}
%Then the acceleration of the UAV can be expressed by:

%where $T$ is the thrust applied to the UAV, which is considered proportional to the rotational speed of the rotors of the UAV: 
%\begin{equation}
%T =  k_{T}\sum_{i=1}^{4} \omega_{i}^{2}    
%\end{equation} where $k_{T}$ is a constant parameter.}

Although it is possible for a human to directly control the acceleration of a robot, acceleration control is not typically implemented because it is not intuitive for the human operator. Instead, it is more common for the human to issue commands related to either the position of the robot or the velocity of the robot \cite{Niemeyer2008}. The human's position and velocity commands can be compared to the robot's current state to calculate the acceleration that the human is commanding. 

For example, in the implementation tested in Section \ref{sec:user study}, we use a force-feedback haptic device as the control interface that the user manipulates to control the UAV. We implement a rate-control scheme, in which the position of the control interface, $\vec{p}_{i}$, is scaled by constant factor, $K_{v}$, to generate a velocity command, $\vec{v}_{c}$: 
\begin{equation} \label{eq:standardRate}
\vec{v}_{c} = K_{v}\vec{p}_{i}.
\end{equation}

The user's acceleration command can then be calculated by: 
\begin{equation} 
\vec{u}_{ref} = \frac{{\vec{v}}_{c}-\vec{v}_{UAV}}{dt},
\end{equation}
where $\vec{v}_{UAV}$ is the current velocity of the UAV and $dt$ is the rate of the control loop. 

Given a human operator's input $\vec{u}_{ref}$ for the control system, we use a CBF to find a safe control input, $\vec{u}$, that is closest to the human operator's input. Details of our CBF implementation is given below in Section \ref{sec:CBF}.

We then calculate a force,  $\vec{F}$, which is applied to the user through a force-feedback control interface:
\begin{equation} 
\vec{F} = K_{f} (\vec{u} - \vec{u}_{ref})
\end{equation}
where $K_{f}$ is a constant parameter to adjust the magnitude of the haptic feedback. The haptic feedback is proportional to the difference between the human control input and the safe control input calculated by CBF, which means there is no haptic feedback when the human operator issues a safe command. Haptic feedback will be generated when the human control input violates the safety constraints of the CBF. We note that violating the safety constraints of the UAV does not necessarily imply that the UAV will crash during the next iteration of the control loop. Instead, the safety constraints can be designed conservatively, so that the UAV will crash only after the same control input is applied over an extended duration. This property enables us to design conservative safety constraints that will allow human operator time to react to the haptic feedback.

\subsection{Control Barrier Function Implementation} \label{sec:CBF}
In this section we provide a brief introduction to CBFs, and  refer the reader to \cite{Ames2019,Ames2014,Nguyen2016,Xiao2019} for complete details. CBFs can be applied to continuous time affine control systems of the form:
\begin{equation}
\dot { \textbf{x} } = f ( \textbf{x} ) + g ( \textbf{x} ) \textbf{u}
\end{equation}
with $f$ and $g$ locally Lipschitz, $ \textbf{x}\in \mathbb { R } ^ { n }$ and $\textbf{u}\in \mathbb { R } ^ { m }$. As a reminder, CBFs ensure that if the control system is in a safe state, the control input, $\textbf{u}$, applied to the system will keep the system in a safe state. 

Barrier functions are used to define the set of all safe states, $C$, as follows: 

\begin{equation}
C : = \left\{ \textbf{x} \in \mathbb { R } ^ { n } : b ( \textbf{x} ) \geq 0 \right\}
\end{equation}
A barrier function must be a continuously differential function with respect to time, for which there is an extended class $ {K} _ {\infty} $ function $\alpha$ such that, for the given control system: 

% \begin{equation}
% L _ { f } b ( x ) + L _ { g } b ( x ) u + \alpha ( b ( x ) ) \geq 0
% \end{equation}
\begin{equation}
\sup _{\textbf{u} \in U}\left[L_{f} b(\textbf{x})+L_{g} b(\textbf{x}) \textbf{u}\right] + \alpha(b(\textbf{x})) \geq 0
\end{equation}
for all $\textbf{x} \in C$. Here, $L_{f}$ and $L_{g}$ are the Lie derivatives of function $b$ to $f$ and $g$ respectively.

In the domain of robotics, if the state vector $\textbf{x}$ only includes positions coordinates x, y, and their time derivatives, then the function b(\textbf{x}) could help define the physical region of safe space. For example, if there is a circular obstacle centered at the origin with radius $r$, then the safe set states would be all positions outside this circle. In this case, $b$ can be described as 

\begin{equation}\label{eqn:barrier}
b(\textbf{x}) = x_i^2 + y_i^2 - r^2
\end{equation}
which would mean that the set of all safe states, C, is 
\begin{equation}
C : = \left\{ \textbf{x} \in \mathbb { R } ^ { n } : x_i^2 + y_i^2 - r^2 \geq 0 \right\}
\end{equation}

Barrier functions are used to design safety constraints for the optimization problem defined in (\ref{eqn:opimization}). When designing such safety constraints for dynamic systems, it is advantageous to include higher-order derivatives of the barrier function. For example, the barrier function described in (\ref{eqn:barrier}), only considers the position of the robot. However, the velocity and acceleration with which the robot approaches the restricted region should matter because a fast approach towards the object is less safe than a slow approach.  Nguyen and Sreenath and Xiao and Belta developed methods to include higher-order derivatives of the barrier function in the safety constraint while guaranteeing safety \cite{Nguyen2016,Xiao2019}. Following these methods, in this paper we implement safety constraints of the form: 
\begin{equation} \label{eqn:safetyConstraint}
    \ddot{b} (\textbf{x})+ 2 p \dot { b } (\textbf{x}) +p ^ { 2 } b (\textbf{x}) \geq 0
\end{equation}
where the barrier function $b$ describes a physical region of safe space and $p$ is a constant parameter that must be greater than $0$. A larger value of $p$ leads to a more conservative constraint, meaning that a constraint with a higher $p$ be violated for a given control input, even when a constraint with a lower $p$ value is satisfied. We note that when $\dot{\textbf{x}}$ and $\ddot{\textbf{x}}$ are zero, this higher order constraint reduces to $b ( \textbf{x} ) \geq 0$. This means that a robot can get arbitrarily close to the physical boundary if it moves at very low speeds. This property may be useful when designing haptic feedback, because hovering near an obstacle will not violate a safety constraint, and thus will not result in a force being exerted on the user.

\section{USER STUDY DESIGN}\label{sec:user study}

We conducted a user study to evaluate the effects force feedback generated through CBFs on human performance in a simulated UAV teleoperation task. This study was approved as exempt by the Boston University Institutional Review Board under protocol number 5070E. 

\subsection{Experimental Setup}\label{exp_setup}
\begin{figure}[t] 
      \centering
      \includegraphics[width=0.9\columnwidth, trim={0cm 0cm 0 0cm},clip]{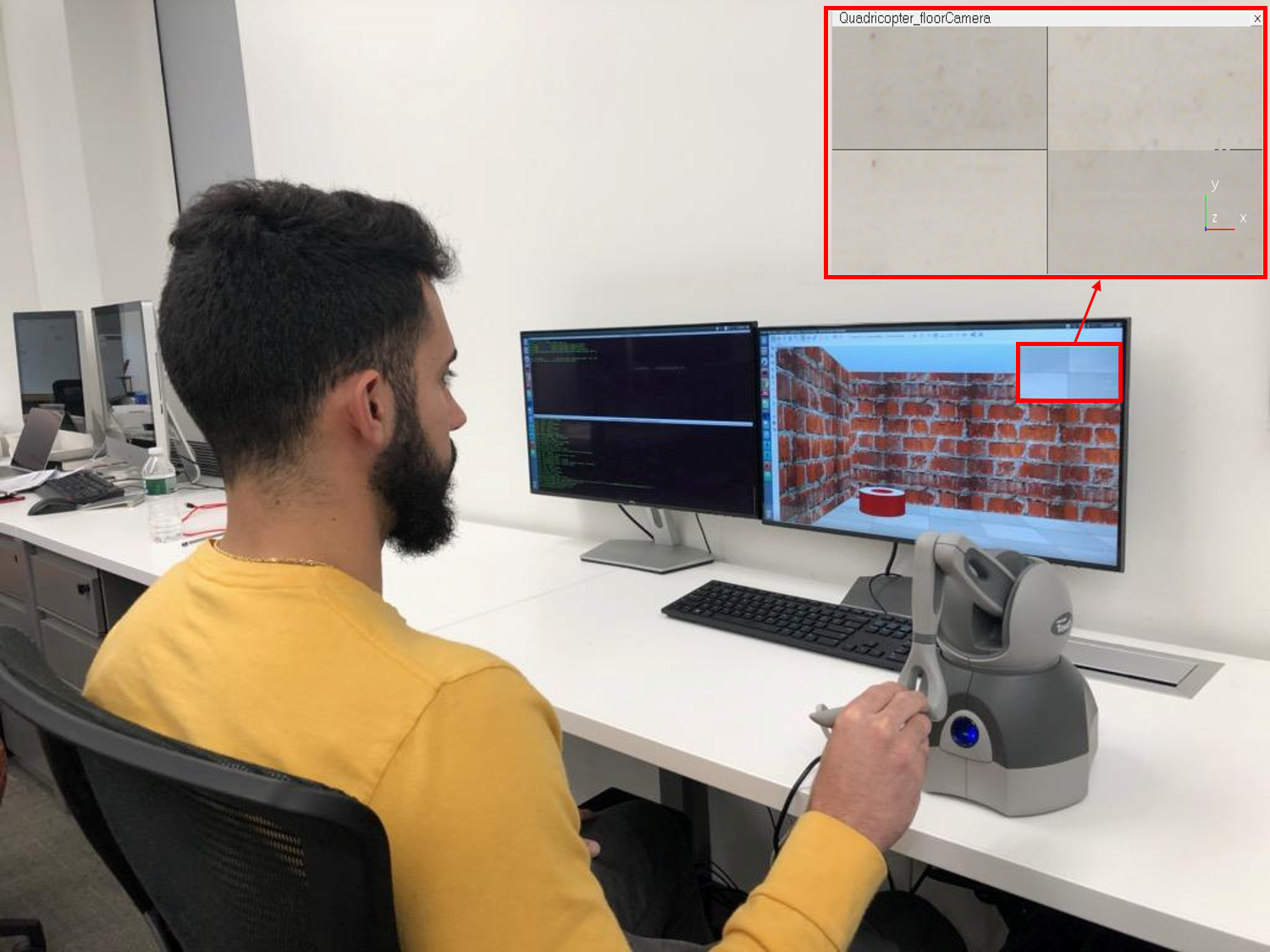}
      \caption{\small Each participant used a haptic joystick to control the UAV in a simulated hallway environment. Participants were shown a first-person view provided by a forward-facing camera on the UAV. Participants were also shown the view captured by a downward-facing camera in a smaller window.} 
      \label{fig:setup}
    %   \vspace{-15pt}
\end{figure}

As shown in Fig. \ref{fig:setup}, each subject controlled the motion of a virtual quadrotor UAV in a simulated 3D environment. The virtual environment and UAV were simulated using robot simulator V-REP \cite{rohmer2013v}. During the study, the human operator controlled the UAV through a virtual hallway that contained four turns and five targets. The subject had to ``inspect'' each target by flying the UAV directly over the target. An overhead view of the environment and the five targets is shown in Fig. \ref{fig:scene}. The simulated robot had a radius of $0.25 m$ and the width of the hallway was $2$ m.

Each subject used a 3D Systems Touch Haptic Device to control the motion of the robot. As shown in Fig \ref{fig:setup}, participants viewed the simulated environment through a forward-facing camera and a bottom-facing camera. The view provided by the forward-facing camera was displayed on a 24-inch computer monitor. The view provided by the bottom-facing camera was shown as a 5.5-inch insert at the top right of the screen. 

In this experiment, the height of the robot above the floor was fixed and the user only had control of the robot's horizontal position and yaw angle. A virtual spring was used to constrain the haptic device to a plane parallel to the tabletop surface. The orientation of the UAV was controlled using the two buttons on the stylus of the haptic device, with one button commanding a counterclockwise rotation and the other commanding a clockwise rotation. The rate of rotation of the UAV was 0.038 rad/s when a button was pressed. 

A velocity control scheme was implemented so that the user's displacement of the joystick was mapped to the UAV's commanded velocity. A light-spring was implemented to help the user return the haptic joystick to the center. Furthermore, a dead-zone of 1 cm was implemented to help enable the user to hover the UAV. The distance of the stylus from the dead-zone was mapped to the UAV's commanded velocity through a constant of 2 $\frac{\textrm{cm}}{m/s}$. The user controlled the robot from a first-person perspective, so that moving the stylus forward (i.e. away from the user) would result in a motion in the direction of the UAV's front-facing camera, moving the stylus to the left would result in a motion command directly to the UAV camera's left, and so on.

%The orientation of the front camera (i.e. the desired yaw angle of the UAV) was controlled by the buttons on the stylus of the Geomagic Touch. The simulated walls would constrain the robot's position, meaning that the person could not command the position of the robot to penetrate the virtual wall. 

%The position of the joystick is measured in terms of its maximum displacement of the stylus. Robot Operating System (ROS) \cite{ROS} was used to subscribe the control command from the haptic joystick. The robot simulator V-REP with Python API was employed to create the virtual environment. 
%Participants used the joystick with their dominant hand to navigate the UAV quadrotor to each of the target locations through the FOV of the front camera. Subjects were also provided a FOV of the bottom camera pointing to the floor of the simulated environment.

\subsection{Evaluated Methods}
Each subject tested the following three haptic feedback conditions methods:
\begin{itemize}
\item \textbf{N} : No haptic feedback is provided to the user. Note: in this condition, the virtual springs are still used to constrain the haptic device to the horizontal plane and help the user recenter the stylus. 
\item \textbf{PRF} : Haptic feedback is generated using the parametric risk field (PRF), proposed in \cite{Lam2009}. 
\item \textbf{CBF} : Haptic feedback is generated using the control barrier functions, as newly proposed in this paper. \end{itemize}

\subsubsection{PRF Haptic Feedback Implementation} 
As described in \cite{Lam2009}, a PRF calculates the risk of collision between the UAV and its environment. A PRF first finds the region around the UAV, termed the critical region, so that if an obstacle was within this region the UAV would unavoidably crash with the obstacle. PRF then defines a boundary around the critical region. The width of this boundary region is $d_{0}$ which is chosen so that if an obstacle is located at or beyond the outer edge of the boundary it would pose no threat of collision. Therefore, a risk of collision is only calculated for obstacles in the boundary region. The risk of collision for each obstacle is determined by the function  by some risk function, $p (\frac{d}{d_{0}})$, where $d$ is the distance between the obstacle and the critical region. As recommended in \cite{Lam2009}, we chose a risk function to be:
\begin{equation}
p \left( \frac { d } { d _ { 0 } } \right) = \cos \left( \frac { d } { d _ { 0 } } \frac { \pi } { 2 } + \frac { \pi } { 2 } \right) + 1
\end{equation}

In our implementation we set to be $d_o = 0.5 m$. The magnitude of the force generated by the haptic device is given by $M = K_{\textrm{PRF}}p$, where $K_{PRF}$ is chosen so that a risk factor of 1 would correspond to the maximum force output of the haptic device, which is 3.3 N. The direction of the force feedback points directly away from the obstacle posing the highest risk of collision. 
%The critical region depends on the robot’s current velocity an maximum acceleration

\subsubsection{CBF Haptic Feedback Implementation}
The unsafe region of this environment would be any position of the UAV's center of mass, $x, y$ that results in a collision between the UAV and the wall. In the environment used in this experiment, this is equivalent to staying within the bounds of the outer rectangle of the environment and staying out of the bounds of the three smaller rectangles within this environment. 

To enforce that the UAV stays within the bounds of the outer rectangle, we write one barrier function for each of the four walls. These barrier functions simply take the form of the line to which each wall is constrained, using a sign convention so that the UAV will be constrained to the proper side of the wall (i.e. to the left of the rightmost wall). We then create a safety constraint, of the form presented in (\ref{eqn:safetyConstraint}), for each of the four barrier functions. 

To enforce that the UAV stays outside the bounds of the smaller rectangles, we applied an extended version of the super-ellipsoid function \cite{barr1981}, to describe the region of space contained by the region of space defined by the Minkowski sum of the UAV and the rectangle. The equation for this super-ellipsoid takes the form: 

\begin{equation}
\left(\frac{|x|}{a}\right)^{2 a / r}+\left(\frac{|y|}{b}\right)^{2 b / r} =1
\end{equation}
where $a$ is the length of the rectangle plus the diameter of the UAV, $b$ is the width of the rectangle plus the diameter of the UAV, and $r$ is the radius of the UAV. 

Then the barrier function for a super-ellipsoid is then given by:
\begin{equation}
b = \left(\frac{|x|}{a}\right)^{2 a / r}+\left(\frac{|y|}{b}\right)^{2 b / r} -1 
\end{equation}

%To define the region of space within any polygon 

%A simulated environment, shown in Fig. \ref{fig:scene}, was used in this study. It contains a constrained hallway with four turns. The width of the hallway is $2$ m. There are four goal locations along with the hallway and one final target location at the end of the hallway. \textcolor{blue}{
%function, which is proposed in \cite{barr1981}, to solve this problem and the form of the equation is described as follows:
%\begin{equation}
%\left(\frac{|x|}{a}\right)^{2 a / r}+\left(\frac{|y|}{b}\right)^{2 b / r}+ %\left(\frac{|z|}{c}\right)^{2 c / r} =1
%\end{equation}
%where $a, b, c$ are the length, width and height of the obstacles respectively and $r$ is the rounding radius of the approximated obstacle.}

We use a Euclidean norm to determine the distance between the human's provided control input $u_{ref}$ and the safe-input returned by the CBF. Therefore, our CBF can be described as: 
\begin{equation}
\begin{array} { c l } { \vec{u}  = \underset { \vec{u} \in \mathbb { R } ^ { m } } { \operatorname { argmin } } } & { \frac { 1 } { 2 } \| \vec{u} - \vec{u}_{ref}\| ^ { 2 } \quad \quad}  \\  \text { s.t. }  &  \ddot{b_i} (\textbf{x})+ 2 p \dot { b_i } (\textbf{x}) +p ^ { 2 } b_i (\textbf{x}) \geq 0 \quad \forall i = 1:7 \end{array}
\end{equation}
where the first four barrier functions, $b_i$, are generated from the requirement to stay within the walls of the outer rectangle, and last three barrier functions are generated from the requirement to say outside the bounds of the three inner rectangles. Using a Euclidean norm, the optimization problem of the CBF becomes a Quadratic Program, which we solved using the Python interface for the Gurobi Optimizer.

%where a Quadratic Program (QP) based controller is used to find a safe control input, $\vec{u}$, that is closest to the human operator's input.
%Since the teleoperation of UAVs aims to protect the UAV from colliding with the obstacles, the design of CBFs becomes designing a group of barrier functions which are used to approximate the physical shape of the obstacles. A detailed description of designing the functions is given in Section \ref{exp_setup}.

\subsection{Experimental Procedure}

\begin{figure}[t] 
  \centering
  \includegraphics[width=0.75\columnwidth, trim={0cm 0cm 0 0cm},clip]{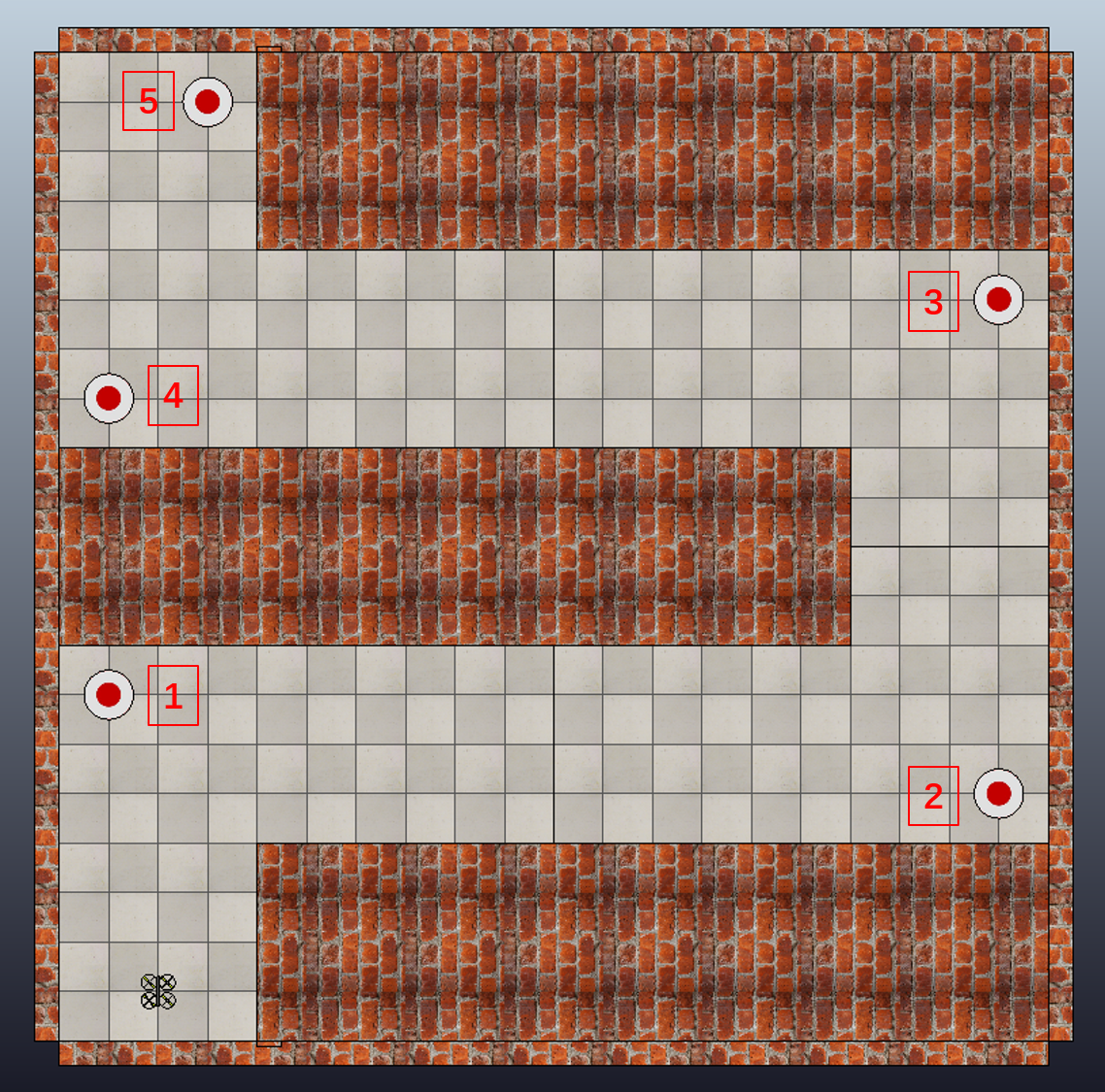}
  \caption{\small Overhead view of the simulated 3D hallway. The numbers indicate the sequence of the targets that each subject was asked to ``inspect". The UAV is shown at its starting location at the bottom left of the figure.}
  \label{fig:scene}
  \vspace{-15pt}
\end{figure}

Twelve subjects participated in this user study (six females, aged 20-33, two left-handed). Participants held the stylus of the haptic with their dominant hand and the haptic device was positioned to align with the subject's corresponding shoulder. 

A within-subject experimental protocol was used. Each subject completed a block 8 trials with each of the three control methods. The presentation order of the control methods was counterbalanced using a Latin Square.  During each trial, the subject flew the UAV along the hallway of the simulated environment, while ``inspecting'' each of the 5 targets shown in Fig.\ \ref{fig:scene} in order. Targets were inspected by flying the UAV directly over each target. For each trial, the UAV began in the position shown in Fig.\ \ref{fig:scene}, with the camera pointing directly along the hallway. A trial began when the participant issued the first non-zero velocity command to the UAV. The trial ended when either the participant maneuvered the UAV over the final target location or when the participant crashed the UAV into the wall in such a way that the simulated UAV lost flight. Participants were told to complete the task as quickly as possible, without colliding the quadrotor with the simulated walls in the experiment. 

After completing the block 8 trials for each method, subjects provided subjective measures of their experience using the NASA Task Load Index (NASA-TLX) \cite{TLX}. %After all three control methods were tested, subjects completed a final survey of NASA-TLX.

\begin{figure*}[t] 
\centering
\includegraphics[width=7in,trim={0cm 0cm 0 0cm},clip]{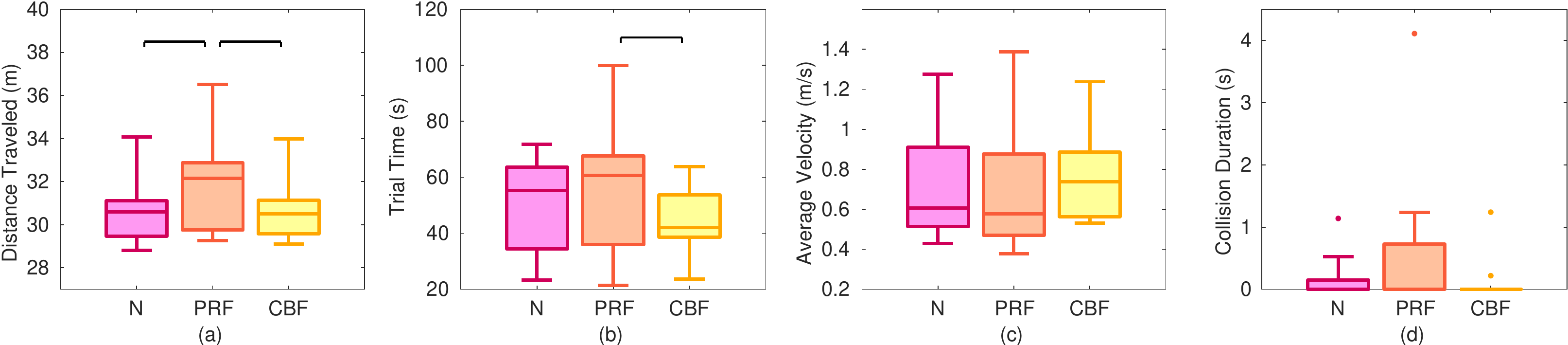}
\caption{\small Results of the user study as measured by  (a) total distance traveled by the UAV, (b) trial time,  (c) average velocity, and (d) duration of collision. Black brackets indicate significance between conditions at the p$<$0.05 level.}
\label{fig:metricResults}
\end{figure*}

\begin{figure*}[t] 
\centering
\includegraphics[width=7in,trim={0cm 0cm 0 0cm},clip]{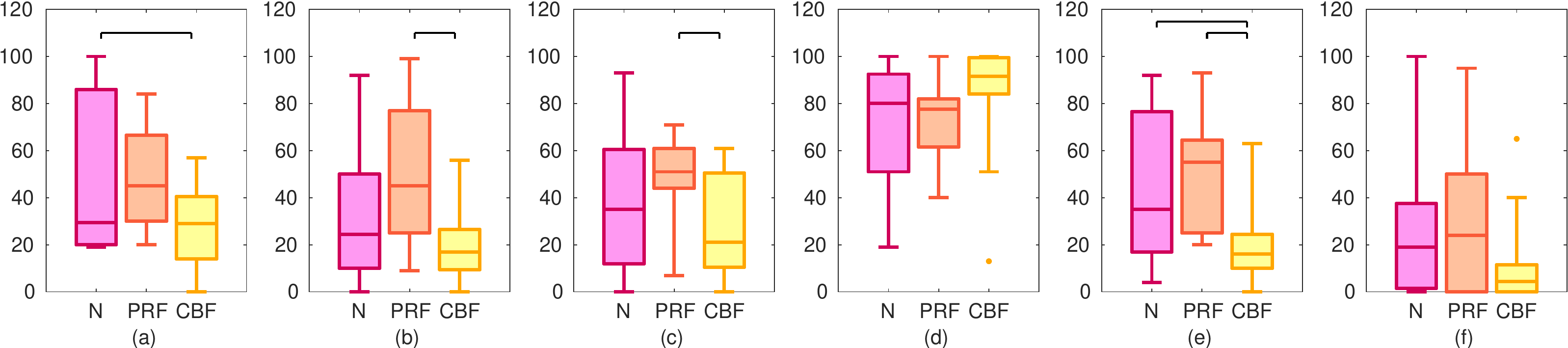}
\caption{\small  Results of the NASA-TLX survey as measured by (a) Mental Demand, (b) Physical Demand, (c) Temporal Demand, (d) Performance, (e) Effort, and (f) Frustration. Black brackets indicate significance between conditions at the p$<$0.05 level.}
\label{fig:resultsTLX}
\vspace{-15pt}
\end{figure*}

\subsection{Measures and Analysis}
 We recorded the UAV's position at a rate of 20Hz. We recorded the number of successful trials, which ended when the UAV reached the final target, and the number of failed trials, which ended when the UAV crashed into the wall such that it lost flight. To examine the effect of the condition of failure rate, we fit a Poisson regression model with the number of failures at the outcome, testing condition as the co-variate of interest, and subject as a fixed effect.
 
 For each successful trial, we compute the following metrics to evaluate the user's performance when using each of the three methods:
\begin{itemize}
\item $D_{total}$: Total distance traveled by the UAV during the trial. A small value of $D_{total}$ means a better economy of motion.
\item $T_{trial}$: The time difference between the start and end of a trial. A smaller $T_{trial}$ implies a better performance in speed.
\item  $V_{avg}$: The average linear velocity of the virtual UAV. A greater value of  $V_{avg}$ indicates a better performance.
\item $T_{collision}$: It was possible for the UAV to lightly contact the simulated wall and remain in stable flight. $T_{collision}$ is the total duration of time that the simulated robot was in contact with the simulated walls. A small value of $T_{collision}$ is preferable.

% \item $N_{fail}$: The number of the trials for each subject that fails to finish the task. A smaller value of $N_{fail}$ means better performance in safety.
% and ranked each of the methods according to: 
% \begin{itemize}
% \item their favorite method
% \item the method they would choose to have a better situational awareness
% \item the method they would choose to accomplish a task safely.
% \item the method they would choose to accomplish a task quickly.
% \item the method which is most mentally taxing
% \item the method which is most physically taxing
% \end{itemize}
\end{itemize}

These four metrics were averaged over the successful trials completed by each subject for each method. Repeated measures analysis of variance (rANOVA) was used to determine whether the method had any effect on task performance. When a significant difference in subject performance was found, Tukey's test was performed at a confidence level of $\alpha = 0.05$ to determine which methods led to significant differences in the metric. %All data analysis was performed using MATLAB's built-in statistical functions. 

%However, the results of the four metrics were only obtained from the data of the successful trials, which means that if the trial is failed, the data will not be recorded. Therefore, we also recorded the number of trials for each subject that fails to finish the task. 

\section{RESULTS}\label{sec:results}

\begin{figure}[t] 
  \centering
  \includegraphics[width=0.8\columnwidth, trim={0cm 0cm 0 0cm},clip]{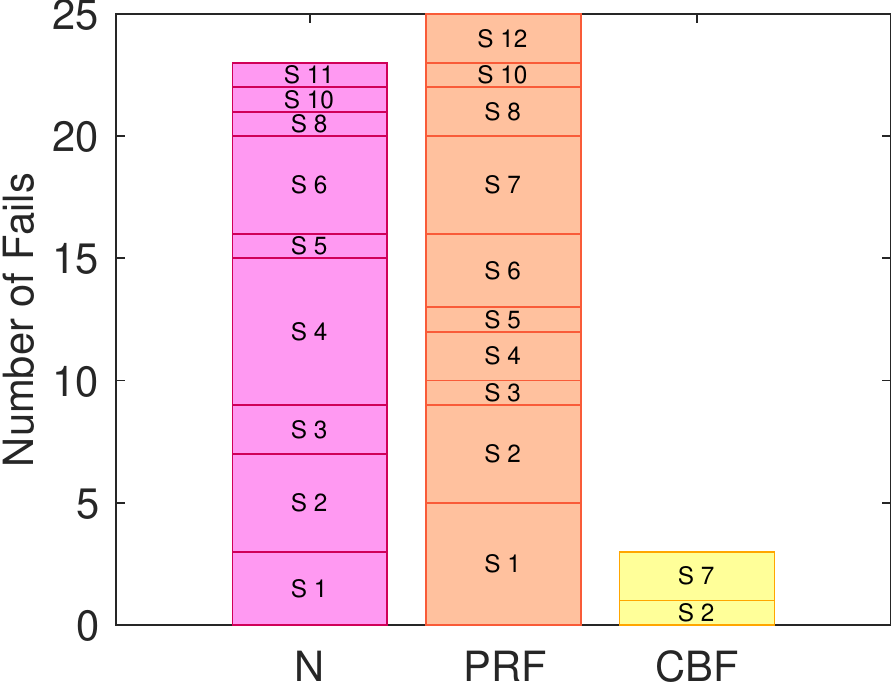}
  \caption{\small Results of the user study as measured by the number of failed trials.  Segments of each bar indicate the number of failures of individual subjects, as labeled.}
  \label{fig:number_fail}
  \vspace{-15pt}
\end{figure}

The results of the failure rate of each subject for each method are shown in Fig. \ref{fig:number_fail}. There were 23 total failed trials under the no haptics condition, 25 failed trials under the PRF condition, and 3 failed trials under the CBF condition. The CBF is associated with 18\% of the failure rate as the no haptics condition (95\% confidence interval (CI):  6.7\%-54.1\%, p $<0.001$) and 16\% of the failure rate as the PRF condition (95\% CI: 5.59\%-46\%, p $<0.001$) 

%p = 0.0009  p = 0.0009. 

%PRF is 107\% of No haptics (95\% 41\%-338\%) p = 0.773
%CBF is 16\% of PRF (95\% 5.59\%-46) p = 0.0009
%CI:  of the failure rate of the no haptic condition and 

As shown in Fig. \ref{fig:metricResults} (a), the haptic feedback condition has a significant effect on the total distance traveled by the simulated UAV quadrotor (F(2,22)=9.15,  p = 0.0013). The distance traveled by the quadrotor when under the PRF condition, was significantly longer than the distance traveled under Conditions N and CBF. There is no significant difference in $D_{total}$ when comparing Condition N against Condition CBF.  

When considering the results of the trial duration, the haptic feedback condition also has a significant effect (F(2,22) = 6.34, p = 0.0067). As shown in Fig. \ref{fig:metricResults} (b), participants took significantly shorter to complete the task with CBF-generated haptic feedback, as compared to PRF-generated haptic feedback. However, no significant differences were found when comparing no haptic feedback against PRF and CBF haptic feedback.

The results did not show significant difference among three tested methods, in terms of average velocity (F(2,22) = 1.97, p = 0.16) and collision duration (F(2,22) = 1.34, p = 0.28) for the successfully completed trials. 

The subjects' subjective rating of workload and task performance, measured the NASA-TLX, are shown in Fig. \ref{fig:resultsTLX}.  No significant differences were found when comparing the effect of haptic feedback type on ratings of performance (F(2,22) = 0.99,  p = 0.39) and frustration (F(2,22) = 2.40,  p = 0.11). Significant differences were found when comparing the effect of haptic feedback type on mental demand (F(2,22) = 4.84,  p = 0.018), physical demand (F(2,22) = 7.83,  p = 0.0027), temporal demand (F(2,22) = 4.42,  p = 0.024), and effort (F(2,22) = 7.20,  p = 0.0039). 
Condition CBF has significantly lower mental demand compared with Condition N. Both physical demand and temporal demand are lower for Condition CBF as compared with the other methods. Participants reported higher levels of effort when using Condition N and Condition PRF, as compared to Condition CBF.

\section{Discussions}\label{sec:discussion}
The reduced number of failed trials indicates that Conditions CBF resulted in the best safety performance when compared with other tested methods. This indicates that CBF-based haptic feedback can help the operator issue safe commands to a UAV.  Although more investigation is needed, this result may also indicate that CBF-based haptic feedback can lead to higher situational awareness for operators. Both of these interpretations are supported by the fact that subjects rated mental demand and effort significantly lower when using CBF haptic feedback.

The results of the user study indicate that safety, efficiency, and user task performance are all improved with CBF haptic feedback and compared to PRF haptic feedback. We note that there are two key differences between these methods. First, CBF haptic feedback is designed to help the user issue a safe control input that is nearest to their current command, while PRF haptic feedback will help guide the user's hand directly from the object with the highest risk. Second, when the UAV is near an obstacle CBF will only generate haptic feedback is the UAV is actively approaching the obstacle with an unsafe velocity and acceleration, while PRF will always generate a repulsive force. Further research is needed to understand the effects of each of these differences. 

%Furthermore, PRF maybe has unhelpful haptic feedback when it is used to complete the task (longer traveled distance than no haptics). Comparison of PRF and CBF shows that it is likely beneficial to design haptic feedback to guide the user to a safe state, as defined by barrier functions. 

Among successful trials, no difference was found between CBF and N. This indicates that in a successful trial, CBF doesn't sacrifice the task efficiency as measured by the distance traveled by the robot, trial time, and average UAV velocity. However, these interpretations of these results must take into account that there are many fewer failed trials with CBF. %Furthermore,  the NASA-TLX showed 

%, which means no significant extra workload was caused by adding CBF-based haptic feedback, as also indicated by the similar physical and temporal demands between CBF and N. However these interpretations of these results must take into account that there are many fewer failed trials with CBF, which indicates that, at the same time, CBF has significant contribution with respect to collision avoidance.

%The significantly longer distance traveled by the UAV quadrotor under the PRF-based method indicates that some extra movements were caused by the unnecessary force feedback generated by PRF. Under the CBF-based method, the force feedback is provided when the user command the robot to an unsafe state, while the PRF-based method are providing the force feedback when there is risk of collision. We believe that it's because the risk of collision for the PRF is still position-based, even though the velocity and acceleration of the UAV are taken into account when building the shape potential field. Therefore, we believe that only providing the haptic feedback when it is necessary can help reduce the workload in the teleoperation of the UAV quadrotor.

%Even though no significant results were shown in the metric of average velocity, the results of the Mental Demand of the TLX survey still show the cues that the human operator has more confidence when controlling the UAV quadrotor under Mehthod CBF , as also indicated by the fact that 
 
\section{CONCLUSIONS AND FUTURE WORK}\label{sec:conclusion} 

In this paper, we present a novel CBF approach to generate haptic feedback for human operators teleoperating UAVs. The result of a user study comparing the effects CBF haptic feedback to no haptic feedback and state-of-the-art PRF haptic feedback shows that generating haptic feedback using CBFs is a promising approach. In the future, we will further investigate which aspects of CBF feedback are most helpful in the teleoperation of UAVs. We will also test our methods 3-dimensional environment using a real quadrotor. Finally, the current implementation is tested with an environment that is known \textit{a priori}, in the future, we will work to generate CBF haptic feedback using sensor data from real UAVs.

\section*{ACKNOWLEDGMENT}
We would like to thank Wei Xiao for the insight into high order control barrier functions.

\bibliographystyle{IEEEtran}
\bibliography{papers}

\begin{thebibliography}{10}
\providecommand{\url}[1]{#1}
\csname url@samestyle\endcsname
\providecommand{\newblock}{\relax}
\providecommand{\bibinfo}[2]{#2}
\providecommand{\BIBentrySTDinterwordspacing}{\spaceskip=0pt\relax}
\providecommand{\BIBentryALTinterwordstretchfactor}{4}
\providecommand{\BIBentryALTinterwordspacing}{\spaceskip=\fontdimen2\font plus
\BIBentryALTinterwordstretchfactor\fontdimen3\font minus
  \fontdimen4\font\relax}
\providecommand{\BIBforeignlanguage}[2]{{%
\expandafter\ifx\csname l@#1\endcsname\relax
\typeout{** WARNING: IEEEtran.bst: No hyphenation pattern has been}%
\typeout{** loaded for the language `#1'. Using the pattern for}%
\typeout{** the default language instead.}%
\else
\language=\csname l@#1\endcsname
\fi
#2}}
\providecommand{\BIBdecl}{\relax}
\BIBdecl

\bibitem{mccarley2005human}
J.~S. McCarley and C.~D. Wickens, ``Human factors implications of uavs in the
  national airspace,'' Aviation Human Factors Division, Savoy, IL, Tech. Rep.,
  2005.

\bibitem{Brandt2010}
A.~M. Brandt and M.~B. Colton, ``Haptic collision avoidance for a remotely
  operated quadrotor {UAV} in indoor environments,'' in \emph{Proc.\
  International Conference on Systems Man and Cybernetics}.\hskip 1em plus
  0.5em minus 0.4em\relax IEEE, 2010, pp. 2724--2731.

\bibitem{Hou2013}
X.~Hou and R.~Mahony, ``Dynamic kinesthetic boundary for haptic teleoperation
  of aerial robotic vehicles,'' in \emph{Proc.\ International Conference on
  Intelligent Robots and Systems}.\hskip 1em plus 0.5em minus 0.4em\relax IEEE,
  2013, pp. 4549--4950.

\bibitem{okamura2009haptic}
A.~M. Okamura, ``Haptic feedback in robot-assisted minimally invasive
  surgery,'' \emph{Current opinion in urology}, vol.~19, no.~1, p. 102, 2009.

\bibitem{pacchierotti2015cutaneous}
C.~Pacchierotti, \emph{Cutaneous haptic feedback in robotic
  teleoperation}.\hskip 1em plus 0.5em minus 0.4em\relax Springer, 2015.

\bibitem{Niemeyer2008}
G.~Niemeyer, C.~Preusche, and G.~Hirzinger, ``Telerobotics,'' in \emph{Springer
  handbook of robotics}.\hskip 1em plus 0.5em minus 0.4em\relax Springer, 2008,
  pp. 741--757.

\bibitem{Omari2013}
S.~Omari, M.-D. Hua, G.~Ducard, and T.~Hamel, ``Bilateral haptic teleoperation
  of {VTOL UAVs},'' in \emph{Proc.\ International Conference on Robotics and
  Automation}.\hskip 1em plus 0.5em minus 0.4em\relax IEEE, 2013, pp.
  2393--2399.

\bibitem{Lam2009}
T.~M. Lam, H.~W. Boschloo, M.~Mulder, and M.~M. Van~Paassen, ``Artificial force
  field for haptic feedback in {UAV} teleoperation,'' \emph{IEEE Transactions
  on Systems, Man, and Cybernetics-Part A: Systems and Humans}, vol.~39, no.~6,
  pp. 1316--1330, 2009.

\bibitem{abbott2007haptic}
J.~J. Abbott, P.~Marayong, and A.~M. Okamura, ``Haptic virtual fixtures for
  robot-assisted manipulation,'' in \emph{Robotics research}.\hskip 1em plus
  0.5em minus 0.4em\relax Springer, 2007, pp. 49--64.

\bibitem{bowyer2013active}
S.~A. Bowyer, B.~L. Davies, and F.~R. y~Baena, ``Active constraints/virtual
  fixtures: A survey,'' \emph{IEEE Transactions on Robotics}, vol.~30, no.~1,
  pp. 138--157, 2013.

\bibitem{okamura2004methods}
A.~M. Okamura, ``Methods for haptic feedback in teleoperated robot-assisted
  surgery,'' \emph{Industrial Robot: An International Journal}, vol.~31, no.~6,
  pp. 499--508, 2004.

\bibitem{prajna2004safety}
S.~Prajna and A.~Jadbabaie, ``Safety verification of hybrid systems using
  barrier certificates,'' in \emph{Proc.\ International Workshop on Hybrid
  Systems: Computation and Control}.\hskip 1em plus 0.5em minus 0.4em\relax
  Springer, 2004, pp. 477--492.

\bibitem{prajna2006barrier}
S.~Prajna, ``Barrier certificates for nonlinear model validation,''
  \emph{Automatica}, vol.~42, no.~1, pp. 117--126, 2006.

\bibitem{Ames2019}
A.~D. Ames, S.~Coogan, M.~Egerstedt, G.~Notomista, K.~Sreenath, and P.~Tabuada,
  ``Control barrier functions: Theory and applications,'' \emph{arXiv preprint
  arXiv:1903.11199}, 2019.

\bibitem{Ames2014}
A.~D. Ames, J.~W. Grizzle, and P.~Tabuada, ``Control barrier function based
  quadratic programs with application to adaptive cruise control,'' in
  \emph{Proc.\ 53rd IEEE Conference on Decision and Control}.\hskip 1em plus
  0.5em minus 0.4em\relax IEEE, 2014, pp. 6271--6278.

\bibitem{Wang2017}
L.~Wang, A.~D. Ames, and M.~Egerstedt, ``Safe certificate-based maneuvers for
  teams of quadrotors using differential flatness,'' in \emph{Proc.\ 2017 IEEE
  International Conference on Robotics and Automation (ICRA)}.\hskip 1em plus
  0.5em minus 0.4em\relax IEEE, 2017, pp. 3293--3298.

\bibitem{Xu2018}
B.~Xu and K.~Sreenath, ``Safe teleoperation of dynamic uavs through control
  barrier functions,'' in \emph{Proc.\ 2018 IEEE International Conference on
  Robotics and Automation (ICRA)}.\hskip 1em plus 0.5em minus 0.4em\relax IEEE,
  2018, pp. 7848--7855.

\bibitem{mellinger2011minimum}
D.~Mellinger and V.~Kumar, ``Minimum snap trajectory generation and control for
  quadrotors,'' in \emph{Proc.\ 2011 IEEE International Conference on Robotics
  and Automation}.\hskip 1em plus 0.5em minus 0.4em\relax IEEE, 2011, pp.
  2520--2525.

\bibitem{voos2009nonlinear}
H.~Voos, ``Nonlinear control of a quadrotor micro-uav using
  feedback-linearization,'' in \emph{Proc.\ 2009 IEEE International Conference
  on Mechatronics}.\hskip 1em plus 0.5em minus 0.4em\relax IEEE, 2009, pp.
  1--6.

\bibitem{Nguyen2016}
Q.~Nguyen and K.~Sreenath, ``Exponential control barrier functions for
  enforcing high relative-degree safety-critical constraints,'' in \emph{Proc.\
  2016 American Control Conference (ACC)}.\hskip 1em plus 0.5em minus
  0.4em\relax IEEE, 2016, pp. 322--328.

\bibitem{Xiao2019}
W.~Xiao and C.~Belta, ``Control barrier functions for systems with high
  relative degree,'' in \emph{Proc.\ 58th IEEE Conference on Decision and
  Control}, 2019, available in arXiv:1903.04706.

\bibitem{rohmer2013v}
E.~Rohmer, S.~P. Singh, and M.~Freese, ``V-rep: A versatile and scalable robot
  simulation framework,'' in \emph{Proc.\ 2013 IEEE/RSJ International
  Conference on Intelligent Robots and Systems}.\hskip 1em plus 0.5em minus
  0.4em\relax IEEE, 2013, pp. 1321--1326.

\bibitem{barr1981}
A.~H. Barr, ``Superquadrics and angle-preserving transformations,'' \emph{IEEE
  Computer graphics and Applications}, vol.~1, no.~1, pp. 11--23, 1981.

\bibitem{TLX}
S.~G. Hart and L.~E. Staveland, ``Development of {NASA-TLX (Task Load Index)}:
  Results of empirical and theoretical research,'' in \emph{Advances in
  psychology}.\hskip 1em plus 0.5em minus 0.4em\relax Elsevier, 1988, vol.~52,
  pp. 139--183.

\end{thebibliography}
\end{document}